\pdfoutput=1

\documentclass[11pt]{article}

\usepackage[]{acl}

\usepackage{times}
\usepackage{latexsym}

\usepackage[T1]{fontenc}

\usepackage[utf8]{inputenc}

\usepackage{microtype}

\usepackage{inconsolata}

\usepackage{algorithm2e}
\usepackage{csquotes}
\usepackage{enumitem}
\usepackage{graphicx}
\usepackage{tabularx}
\usepackage{booktabs}
\usepackage{listings}
\usepackage{amssymb}
\usepackage{subcaption}
\usepackage{geometry}
\usepackage{multirow}
\usepackage{makecell}
\usepackage{array}

\title{On Evaluating the Integration of Reasoning and Action in LLM Agents \\ with Database Question Answering}

\author{
        Linyong Nan$^{1}$
\quad   Ellen Zhang$^{1}$
\quad   Weijin Zou$^{2}$
\quad   Yilun Zhao$^{1}$
\\{\bf
\quad   Wenfei Zhou$^{3}$
\quad   Arman Cohan$^{1,4}$}
\vspace{4pt} \\ 
$^1$Yale University 
\quad $^2$LinkedIn \quad $^3$NVIDIA Corporation \quad $^4$Allen Institute for AI
\\
\texttt{\{linyong.nan, ellen.zhang\}@yale.edu}
}

\begin{document}
\maketitle
\begin{abstract}
This study introduces a new long-form database question answering dataset designed to evaluate how Large Language Models (LLMs) interact with a SQL interpreter. The task necessitates LLMs to strategically generate multiple SQL queries to retrieve sufficient data from a database, to reason with the acquired context, and to synthesize them into a comprehensive analytical narrative. Our findings highlight that this task poses great challenges even for the state-of-the-art \texttt{GPT-4} model. We propose and evaluate two interaction strategies, and provide a fine-grained analysis of the individual stages within the interaction. A key discovery is the identification of two primary bottlenecks hindering effective interaction: the capacity for planning and the ability to generate multiple SQL queries. To address the challenge of accurately assessing answer quality, we introduce a multi-agent evaluation framework that simulates the academic peer-review process, enhancing the precision and reliability of our evaluations. This framework allows for a more nuanced understanding of the strengths and limitations of current LLMs in complex retrieval and reasoning tasks.
\end{abstract}

\section{Introduction}
\label{sec:intro}
\begin{figure*}[t]
    \centering
    \includegraphics[width=\textwidth]{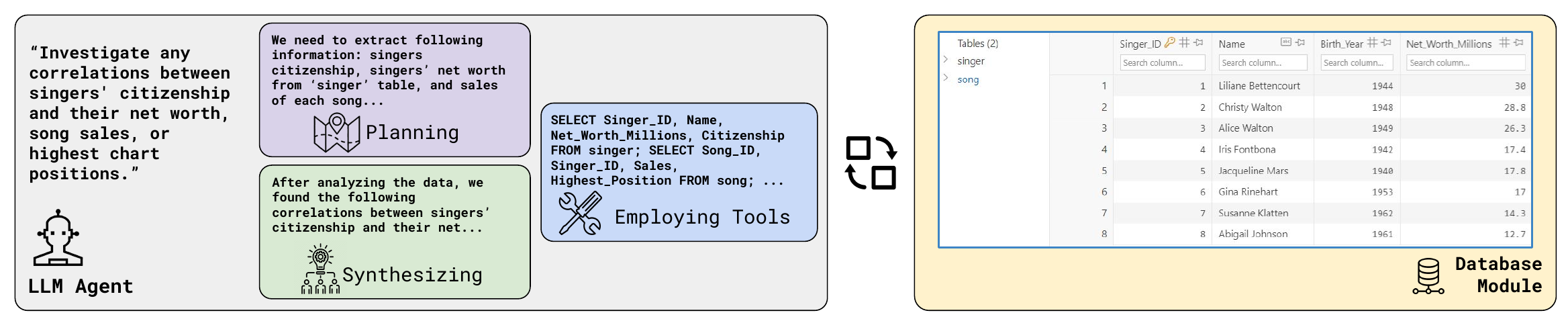}
    \caption{Illustration of our long-form database question answering task. The LLM agent is expected to perform a series of tasks requiring reasoning and actions to interact with the database module.}
    \label{fig:task-illustration}
\end{figure*}

Significant advancements in natural language processing have been driven by the development of Large Language Models (LLMs) \cite{devlin-etal-2019-bert, radford2019language, NEURIPS2020_1457c0d6, chowdhery2022palm, openai2023gpt4}, which have become fundamental components of numerous products used by millions, reshaping people's habits on accessing information. Despite their widespread adoption and impact, LLMs face intrinsic limitations due to their design, including limited context window, stochastic nature 
which makes them less suited for tasks requiring high standards of precision, and extensive computations \cite{mialon2023augmented, 10.1145/3571730, wang2023survey}. Many studies have explored ways to mitigate these constraints by augmenting LLMs with modules/tools of complementary features \cite{nakano2022webgpt, NEURIPS2020_6b493230, lazaridou2022internetaugmented, pmlr-v202-gao23f, parisi2022talm, schick2023toolformer}. In our study, we focus on augmenting LLMs with a symbolic module - a SQL code interpreter - and assess their performance using the long-form database question-answering task that we introduce, illustrated in Figure \ref{fig:task-illustration}. Such augmentation is inevitable for tasks involving databases, as they often far exceed the size of LLMs' context windows\footnote{Enterprise databases can easily store hundreds of millions of records for real-world applications.}, making information retrieval through any means other than SQL inefficient. Additionally, the use of SQL queries brings transparency to the reasoning process of LLM agents, providing a means to validate the accuracy of their generated responses.

LLMs augmented with external modules/tools possess two primary abilities: the capacity to \emph{act}, which involves the use of tools, and the capability to \emph{reason}, which encompasses planning and analyzing the outcomes of actions \cite{mialon2023augmented, madaan2023selfrefine, paul2023refiner, yao2023react, yoran2023answering, shinn2023reflexion}. While numerous studies have evaluated these abilities in different contexts, we contend that some of them focus more on evaluating tool selection and tool employment with less focus on evaluating how LLM agents reflect or synthesize the action results \cite{parisi2022talm, schick2023toolformer, zhuang2023toolqa, li2023apibank}. Other research \cite{shuster2022language, yao2023react, behnamghader2023retrieveraugmented} does examine both the action and reasoning capacities of LLM agents, yet the actions' complexity is not as demanding as in studies with a stronger focus on the action aspect. Our goals are twofold: firstly, to introduce a task that places equal emphasis on the complexities of both action and reasoning, requiring a concerted interaction between them, and secondly, to assess the proficiency of various LLM agents merging these dual aspects into a cohesive performance. Here are our main contributions: 

\begin{itemize}[wide, labelindent=0pt]
\setlength{\itemsep}{0pt}
  \item We introduce a new long-form 
  database question answering task, requiring retrieval, reasoning and synthesis of diverse information from database. We develop a systematic approach for collecting questions, databases, and corresponding answers in a way that ensures the answers are definitive and indisputable, lending greater validity to the evaluation process. The task is challenging in retrieval: on average, it requires the formulation of three SQL queries to gather sufficient information to answer the questions.
  \item We explore the benefits of augmenting LLMs with the SQL code interpreter for our task, by comparing the performance of baseline LLMs given the complete database records but without SQL capabilities against LLM agents that are given database schema and SQL generation capacity. 
  \item In evaluating the performance of agents across all sub-tasks, we identify that planning and tool utilization are the critical challenges in achieving effective coordination. We also delve into the reasons behind their shortcomings. We extend our examination to the generalizability of our results across various LLMs, measuring the disparity in performance between agents using proprietary and open-source LLMs as their foundation. 
  \item Finally, we introduced a multi-agent evaluation framework aimed at enhancing the precision and consistency of the output assessments using \texttt{GPT-4} evaluators.
\end{itemize}

\section{Data Collection}
In constructing our evaluation dataset, we prioritize a robust set of desiderata. These include the intensive retrieval of diverse information from the database, the application of rigorous reasoning over the information retrieved, and the synthesis of facts and inferences into a coherent and comprehensive long-form answer. Our methodology employs a hybrid annotation framework: we leverage the capabilities of \texttt{GPT-4} to generate preliminary annotations, then these annotations are selected and refined through manual intervention to ensure quality and relevance. The specifics and quantitative details of our evaluation dataset are presented in Table \ref{tab:dataset_stats}. We detail the collection of questions in Section \ref{sec:question-generation} and the acquisition of answers in Section \ref{sec:answer-annotation}.

\begin{table}[tb]
\centering
\resizebox{\columnwidth}{!}{%
\begin{tabular}{>{\arraybackslash}m{0.8\columnwidth} >{\centering\arraybackslash}m{0.2\columnwidth}}
\toprule
\textbf{Property} & \textbf{Value} \\
\midrule
\textbf{Evaluation Dataset Size} & 200 \\
\quad - \# Conclusive Questions & 98 \\
\quad - \# Interpretive Questions & 102 \\
\textbf{Reference Answer Length} &  \\
\quad - Conclusive Questions (Avg.) & 132 \\
\quad - Interpretive Questions (Avg.) & 209 \\
\textbf{Database size} &  \\
\quad - \# Tables (Med.) & 4 \\
\quad- \# Columns (Med.) & 4 \\
\quad - \# Data Records (Med.) & 11 \\
\bottomrule
\end{tabular}%
}
\caption{Dataset Statistics. \textbf{Avg.} stands for average and \textbf{Med.} stands for median.}
\label{tab:dataset_stats}
\end{table}

\subsection{Question Generation}
\label{sec:question-generation}
Our starting point is the databases from the Spider dataset \citep{yu-etal-2018-spider}. We introduce a question generation pipeline designed to generate questions and iteratively refine them, addressing common issues encountered during preliminary experiments with \texttt{GPT-4} generated queries. This pipeline can be described as \textbf{Control-Condense-Confirm}, it begins by exerting \textbf{control} over the question generation. We direct \texttt{GPT-4} to generate questions that pertain to specific entities or keywords by using the original questions from the Spider dataset as the basis. These questions are instrumental as they concentrate on distinct column sets from various tables, providing a targeted focus that counters the LLM's propensity to formulate overly broad and indistinct questions. Following the initial control, we often find the questions to be exceedingly detailed. To address this, we \textbf{condense} the content, removing superfluous information. This pruning process not only ensures the questions remain challenging but also leaves room for the model to demonstrate its inferential capabilities. The final phase is the manual review of questions to \textbf{confirm} they are unambiguous and meet all predefined criteria for the task. This step guarantees that the questions are of high quality and align with the specified desiderata of our dataset.

\subsection{Answer Annotation}
\label{sec:answer-annotation}
Building upon the question generation strategy outlined in the previous section, the task of annotating answers to questions is generally an effort-intensive task as it requires the formulation of multiple SQL queries. This task is further complicated by the fact that many databases, such as those in the Spider collection, often contain an insufficient number of data records for a comprehensive answer. We propose a method that employs a \textbf{Conjecture-Construct-Conclude} strategy to circumvent these issues.

The process begins by prompting \texttt{GPT-4} with the question alongside the database schema to \textbf{conjecture} an answer. Subsequently, we \textbf{construct} database records that corroborate this conjectured answer, formatted as \texttt{INSERT} statements. These statements are integrated with the original database's \texttt{CREATE} statements, resulting in a bespoke synthetic database aligned with the question. To ensure the integrity of the synthetic database, we execute the merged statements to confirm the absence of errors and manually inspect the data records' alignment with the conjectured answer. As the final step of our method, we task \texttt{GPT-4} to \textbf{conclude} with a substantiated answer, ensuring that it aligns with the evidences we injected to the synthetic database. This procedure ensures that each question is matched with a definitive answer, backed by verifiable evidence from the database records.

Finally, we examine the question and its corresponding answer. We noticed that a substantial number of questions allow for multiple plausible answers, each subject to interpretation of certain abstract word in the question.\footnote{Such as "impact", "success", "notable trends", etc.} To refine the fairness of evaluations against a reference answer, we categorize all questions as either "Interpretive" or "Conclusive". This classification is based on whether the question demands a clear-cut outcome or allows various valid responses. We provide demonstrative examples in Figure \ref{fig:question-type-demo} of the appendix to illustrate the distinction between these categories. The distribution of questions across these categories is detailed in Table \ref{tab:dataset_stats}.

\section{Methods}
\label{sec:methods}
We aim to evaluate how effective LLMs are at performing a complex task that necessitate working with external modules. We explore five main aspects: (1) the proficiency of LLMs in completing our proposed task through interaction with external modules; (2) the extent of improvement LLMs gain from engaging with external modules; (3) the impact of various interaction strategies on LLM performance and the identification of the most effective one; (4) the challenges that hinder effective interaction; (5) the generalizability of our findings across diverse LLMs and the performance disparities attributed to the usage of different LLMs. We can address the first, second and last aspects by directly evaluating the quality of the final answer generated by LLMs. To delve into the third and fourth aspects, we need to first dissect the "interaction" process within our task into its constituent components.

We propose to decompose the LLMs' expected workflow for our task into three distinct sub-tasks: interaction planning, tool employment, and information synthesis. \textbf{Interaction planning} involves the LLM determining its interaction strategy with the external module, considering the question and past interactions. \textbf{Tool employment} is the phase where the LLM generates module-specific commands for the actual interaction. \textbf{Information synthesis} requires the LLM to review the interaction history and any newly acquired context to compile the key information for the final answer. This framework allows us to refine our second objective into assessing how different compositions of these sub-tasks affect
the quality of the final answer, and also to define the most effective composition. The third objective can be addressed by evaluating LLM's execution within each sub-task.

While the potential configurations of these sub-tasks are vast, this study will narrow its focus to two primary interaction strategies for feasibility:

\begin{itemize}
    \item \textbf{Sequential}: The LLM agent systematically tackles the sub-tasks in a linear, step-by-step fashion, with predetermined sequence: interaction planning, tool employment, and information synthesis. The agent's focus should be on prioritizing both precision and comprehensiveness throughout each juncture.
    \item \textbf{Iterative}: The LLM agent cyclically alternates between interaction planning and tool employment, similar to the self-ask prompting \citep{press2023measuring}. The key aspect of interaction planning in this context is to identify the most crucial information to extract from the database given the previous interactions. The strategy calls for the agent to ensure precision in every single interaction and to achieve comprehensiveness by deciding when to terminate the interaction cycle.
\end{itemize}

Equipped with these strategies, we proceed to empirically explore our central questions.

\section{Experiments}

\subsection{Design}
To prove the key areas identified in Section \ref{sec:methods}, we designed two sets of experiments. The first set evaluates three different types of LLM agents:
\begin{itemize}
    \item No-Interaction: This LLM is tasked with deriving the final answer with a chain-of-thoughts prompting without engaging with the SQL module, i.e. generating SQL queries. To ensure fairness, we supply the complete database records within the prompt for context.
    \item Sequential-Interaction: We implement an LLM agent that utilizes the sequential strategy when working with the SQL module. It begins by devising a plan in natural language to identify the needed information and its sources, proceeds to generate SQL queries to retrieve this information, and concludes by integrating the data into the final answer.
    \item Iterative-Interaction: This strategy employs an LLM agent that iteratively determines the most crucial information to retrieve given the interaction history. The agent articulates this in natural language, crafts the corresponding SQL query, and repeats this process until it elects to stop. The final step involves consolidating the gathered information into a conclusive answer.
\end{itemize}

The second set of experiment focuses on evaluating the generalizability of our findings across various LLMs, as well as comparing their performance. We tested two proprietary LLMs: \texttt{GPT-4} and \texttt{GPT-3.5-turbo}, and six open-source LLMs of different sizes and capabilities: \texttt{Llama-2-[7, 13]b}, \texttt{Code-Llama-[7, 13, 34]b}, and \texttt{Mistral-7b}. The \texttt{Llama-2} models were tested using their chat versions, while the \texttt{Code-llama} and \texttt{Mistral} models were evaluated using versions fine-tuned for instruction-following.

\begin{figure*}
    \centering
    \includegraphics[width=\textwidth]{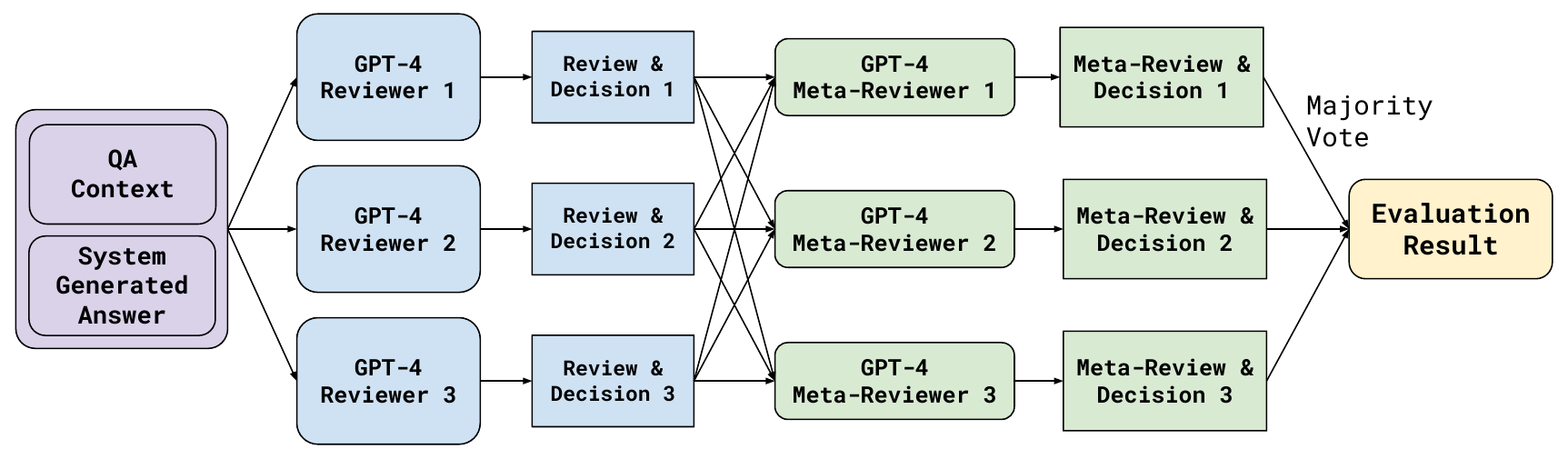}
    \caption{Illustration of our multi-agent evaluation framework. It consists of two tiers of evaluation process.}
    \label{fig:multi-agent-framework}
\end{figure*}

\subsection{Evaluation}
To rigorously assess the performance, we implemented two distinct evaluation methods. Both involve using an LLM for the evaluation process, yet they differ in terms of their reliance on a reference answer. Throughout both evaluation methods, we use \texttt{GPT-4} to ensure consistency.

\subsubsection{Reference-based Evaluation}
In this method, we utilize an LLM to compare the system-generated answer against a reference answer, whose acquirement is detailed in Section \ref{sec:answer-annotation}. The evaluation protocol is adapted to suit the nature of the question: for conclusive questions that demand a specific answer, the LLM evaluator provides a straightforward verdict of either \textit{"match"} or \textit{"not match"} and offers a rationale for its decision. For interpretive questions, which permit a spectrum of answers, the LLM assigns a nuanced score ranging from 1 (no match) to 5 (exact match), reflecting the degree of information overlap with the reference. The scoring rubric for this nuanced evaluation is outlined in Figure \ref{fig:ref-based-rubric} of the appendix.

\subsubsection{Reference-free Evaluation}
Assessing LLM performance on individual sub-tasks is essential, yet the multitude of potential answer pathways complicates the annotation process, making reference-based evaluation impractical. To navigate this challenge, we devised a reference-free evaluation using a \textbf{multi-agent framework} modeled after the academic peer-review system, as illustrated in Figure \ref{fig:multi-agent-framework}. This framework enlists a group of reviewers and meta-reviewers to evaluate the system outputs. Each reviewer receives the question, database schema, and the LLM agent's output for individual sub-tasks. Their role is to critically assess each output across various dimensions and determine if it is \textit{"perfect"} or \textit{"not perfect"}. Meta-reviewers are then presented with the reviewers' assessments and verdicts. Their task is to discern consensus or discrepancies among the reviewers' opinions, evaluate the validity of their critiques, and render a final decision of \textit{"perfect"} or \textit{"not perfect"}. The ultimate evaluation outcome is derived from the majority ruling among the meta-reviewers. To ensure a diversity of perspectives and avoid uniformity in judgment, we configured each \texttt{GPT-4} evaluator with a temperature of 0.7. The specific guidelines used to direct reviewers and meta-reviewers are detailed in Figures \ref{fig:multi-agent-framework-IP-rubric}, \ref{fig:multi-agent-framework-TE-rubric}, \ref{fig:multi-agent-framework-IS-rubric} in the appendix.  

\begin{table*}[h]
\centering
\resizebox{0.9\textwidth}{!}{%
\begin{tabular}{>{\centering\arraybackslash\footnotesize}m{2cm} >{\centering\arraybackslash\footnotesize}m{1.5cm} >{\centering\arraybackslash}m{2cm} >{\centering\arraybackslash}m{2cm} >{\centering\arraybackslash}m{2cm} >{\centering\arraybackslash}m{2cm} >{\centering\arraybackslash}m{2cm}}
\toprule \noalign{\vskip 1mm}
\textbf{LLM} & \textbf{\footnotesize \makecell{Interaction \\ Mode}} & \textbf{\footnotesize \makecell{Match Score \\ (\texttt{C/I})}} & \textbf{\footnotesize \makecell{Plan Length \\ (\texttt{C/I})}} & \textbf{\footnotesize \makecell{\# Generated \\ SQLs (\texttt{C/I})}} & \textbf{\footnotesize \makecell{\# Valid \\ SQLs (\texttt{C/I})}} & \textbf{\footnotesize \makecell{Answer \\ Length (\texttt{C/I})}} \\ \noalign{\vskip 1mm}\toprule \noalign{\vskip 1mm}
\multirow{2}{*}{\footnotesize \texttt{GPT-4}}        & \centering\footnotesize Sequential                & 0.30 / 2.34                                     & 437 / 474                     & 2.96 / 3.28                                & 2.72 / 2.94                               & 197 / 221  \\ 
             & \centering\footnotesize Iterative                 & 0.24 / 2.21                                    & 83 / 101                      & 0.99 / 1.18                                & 0.79 / 0.80                               & 157 / 199  \\ \noalign{\vskip 1mm}\hline\noalign{\vskip 1mm}
\multirow{2}{*}{\footnotesize \texttt{GPT-3.5-turbo}} & \centering\footnotesize Sequential               & 0.28 / 2.04                                    & 297 / 321                    & 2.23 / 2.63                                & 1.84 / 2.30                               & 202 / 193 \\ 
              & \centering\footnotesize Iterative                & 0.15 / 1.49                                     & 223 / 252                     & 1.06 / 1.10                                & 0.89 / 0.81                               & 94 / 90  \\ \noalign{\vskip 1mm}\hline\noalign{\vskip 1mm}
\multirow{2}{*}{\footnotesize \texttt{Llama-2-7b}}   & \centering\footnotesize Sequential                & 0.21 / 1.95                                    & 364 / 347                    & 2.88 / 2.58                               & 1.29 / 1.04                               & 285 / 283  \\ 
             & \centering\footnotesize Iterative                 & 0.23 / 1.88                                    & 63 /92                     & 0.65 /1.03                               & 0.13 / 0.14                              & 252 / 292  \\ \noalign{\vskip 1mm}\hline\noalign{\vskip 1mm}
\multirow{2}{*}{\footnotesize \texttt{Llama-2-13b}}  & \centering\footnotesize Sequential                & 0.06 / 1.41                                    & 398 / 401                    & 3.05 / 1.83                               & 1.32 / 0.86                              & 336 / 339  \\ 
             & \centering\footnotesize Iterative                 & 0.18 / 1.72                                    & 16 / 16                     & 0.20 / 0.28                               & 0.01 / 0.01                              & 310 / 359 \\ \noalign{\vskip 1mm}\hline\noalign{\vskip 1mm}
\multirow{2}{*}{\footnotesize \texttt{Code-llama-7b}} & \centering\footnotesize Sequential               & 0.13 / 1.60                                    & 368 / 390                    & 3.25 / 3.67                               & 1.59 / 1.89                              & 306 / 325  \\ 
              & \centering\footnotesize Iterative                & 0.11 / 1.57                                    & 0 / 0                      & 0 / 0                                  & 0 / 0                                 & 234 / 240 \\ \noalign{\vskip 1mm}\hline\noalign{\vskip 1mm}
\multirow{2}{*}{\footnotesize \texttt{Code-llama-13b}} & \centering\footnotesize Sequential              & 0.17 / 1.62                                    & 389 / 396                    & 4.34 / 5.36                               & 1.95 / 2.78                              & 292 / 314 \\ 
               & \centering\footnotesize Iterative               & 0.15 / 1.51                                    & 115 / 110                     & 0.66 / 1.59                               & 0.26 / 0.70                              & 243 / 245   \\ \noalign{\vskip 1mm}\hline\noalign{\vskip 1mm}
\multirow{2}{*}{\footnotesize \texttt{Code-llama-34b}} & \centering\footnotesize Sequential              & 0.19 / 1.93                                     & 359 / 377                    & 2.79 / 3.31                               & 1.39 / 1.72                              & 314 / 352   \\ 
               & \centering\footnotesize Iterative               & 0.13 / 1.85                                   & 39 / 47                     & 0.4 / 0.33                                & 0.28 / 0.18                              & 248 / 294   \\ \noalign{\vskip 1mm}\hline\noalign{\vskip 1mm}
\multirow{2}{*}{\footnotesize \texttt{Mistral-7b}}     & \centering\footnotesize Sequential              & 0.16 / 1.71                                    & 384 / 379                    & 1.01 / 0.91                               & 0.57 / 0.37                              & 231 / 282 \\ 
               & \centering\footnotesize Iterative               & 0.19 / 1.71                                    & 0 / 0                      & 0 / 0                                  & 0 / 0                                 & 207 / 267  \\ \noalign{\vskip 1mm}\toprule
\end{tabular}%
}
\caption{Reference-based evaluation results and other measurements of the interaction process. \texttt{C} stands for \textbf{conclusive} and \texttt{I} stands for \textbf{interpretive}. Valid SQL indicate SQL queries that are generated by the LLM agent that have non-empty execution results.}
\label{tab:ref-based-full-result}
\end{table*}

\begin{table}[h]
\centering
\resizebox{\columnwidth}{!}{%
\begin{tabular}{>{\centering\arraybackslash}m{0.25\columnwidth} >{\centering\arraybackslash}m{0.23\columnwidth} >{\centering\arraybackslash}m{0.25\columnwidth} >{\centering\arraybackslash}m{0.27\columnwidth}}
\toprule
\textbf{LLM} & \textbf{Interaction Mode} & \textbf{Conclusive Questions Match Rate} & \textbf{Interpretive Questions Match Score} \\
\midrule
\multirow{2}{*}{\footnotesize \texttt{GPT-4}} & No & 0.19 & 2.37 \\
& Sequential & 0.27 & 2.40 \\
\midrule
\multirow{2}{*}{\footnotesize \texttt{GPT-3.5-turbo}} & No & 0.24 & 2.12 \\
& Sequential & 0.30 & 2.13 \\
\bottomrule
\end{tabular}%
}
\caption{LLM agents with vs. without interaction with external SQL modules. Entire database records are provided in the context for LLMs that disable interaction. We sampled 161 out of 200 questions to report their reference-based evaluation results because some databases are too large to fit into the context window of LLMs.}
\label{tab:with-without-tool-results}
\end{table}

\begin{table}[h]
\centering
\resizebox{\columnwidth}{!}{%
\begin{tabular}{>{\centering\arraybackslash}m{0.2\columnwidth} >{\centering\arraybackslash}m{0.2\columnwidth} >{\centering\arraybackslash}m{0.15\columnwidth} >{\centering\arraybackslash}m{0.15\columnwidth} >{\centering\arraybackslash}m{0.15\columnwidth} >{\centering\arraybackslash}m{0.15\columnwidth}}
\toprule
\multirow{3}{*}{\textbf{LLM}} & \multirow{3}{*}{\textbf{Sub-Task}} & \multicolumn{2}{c}{\textbf{Reviewer}} & \multicolumn{2}{c}{\textbf{Meta-Reviewer}} \\
\cmidrule(lr){3-4} \cmidrule(lr){5-6}
& & \textbf{Perf. Rate} & \textbf{Agree.} & \textbf{Perf. Rate} & \textbf{Agree.} \\
\midrule
\multirow{3}{*}{\footnotesize \texttt{GPT-4}} & IP & 0.48 & 0.54 & 0.28 & 0.88 \\
& TE & 0.41 & 0.69 & 0.28 & 0.93 \\
& IS & 0.61 & 0.72 & 0.53 & 0.93 \\
\midrule
\multirow{3}{*}{\footnotesize \texttt{GPT-3.5-turbo}} & IP & 0.14 & 0.80 & 0.09 & 0.96 \\
& TE & 0.18 & 0.82 & 0.13 & 1.00 \\
& IS & 0.26 & 0.79 & 0.22 & 0.95 \\
\midrule
\multirow{3}{*}{\footnotesize \texttt{Llama-2-7b}} & IP & 0.02 & 0.97 & 0.01 & 0.99 \\
& TE & 0.01 & 0.99 & 0.01 & 1.00 \\
& IS & 0.01 & 0.99 & 0.00 & 1.00 \\
\midrule
\multirow{3}{*}{\footnotesize \texttt{Llama-2-13b}} & IP & 0.03 & 0.96 & 0.02 & 0.98 \\
& TE & 0.01 & 0.98 & 0.01 & 0.99 \\
& IS & 0.01 & 0.99 & 0.01 & 0.99 \\
\midrule
\multirow{3}{*}{\footnotesize \texttt{Code-llama-7b}} & IP & 0.02 & 0.93 & 0.01 & 0.99 \\
& TE & 0.01 & 0.98 & 0.01 & 0.99 \\
& IS & 0.02 & 0.98 & 0.01 & 1.00 \\
\midrule
\multirow{3}{*}{\footnotesize \texttt{Code-llama-13b}} & IP & 0.01 & 0.96 & 0.01 & 1.00 \\
& TE & 0.02 & 0.98 & 0.02 & 1.00 \\
& IS & 0.02 & 0.97 & 0.02 & 1.00 \\
\midrule
\multirow{3}{*}{\footnotesize \texttt{Code-llama-34b}} & IP & 0.04 & 0.90 & 0.01 & 0.98 \\
& TE & 0.02 & 0.97 & 0.01 & 0.99 \\
& IS & 0.04 & 0.96 & 0.04 & 0.99 \\
\midrule
\multirow{3}{*}{\footnotesize \texttt{Mistral-7b}} & IP & 0.03 & 0.96 & 0.01 & 0.99 \\
& TE & 0.01 & 0.98 & 0.00 & 1.00 \\
& IS & 0.05 & 0.94 & 0.03 & 0.98 \\
\bottomrule
\end{tabular}%
}
\caption{Reference-free multi-agent evaluation - fine-grained results for different LLMs adopting \textbf{Sequential} interaction strategy. \textbf{IP} stands for Interaction Planning, \textbf{TE} stands for Tool Employment, and \textbf{IS} stands for Information Synthesis. \textbf{Perf. Rate} stands for percentage of instances that (meta-)reviewers considers perfect, and \textbf{Agree.} stands for agreement, and it is calculated with the percentage of instances that (meta-)reviewers reach in unanimous agreement.}
\label{tab:ref-free-full-result}
\end{table}

\begin{figure*}[!ht]
\begin{subfigure}{\columnwidth}
\centering
\resizebox{\columnwidth}{!}{%
\includegraphics[width=\columnwidth]{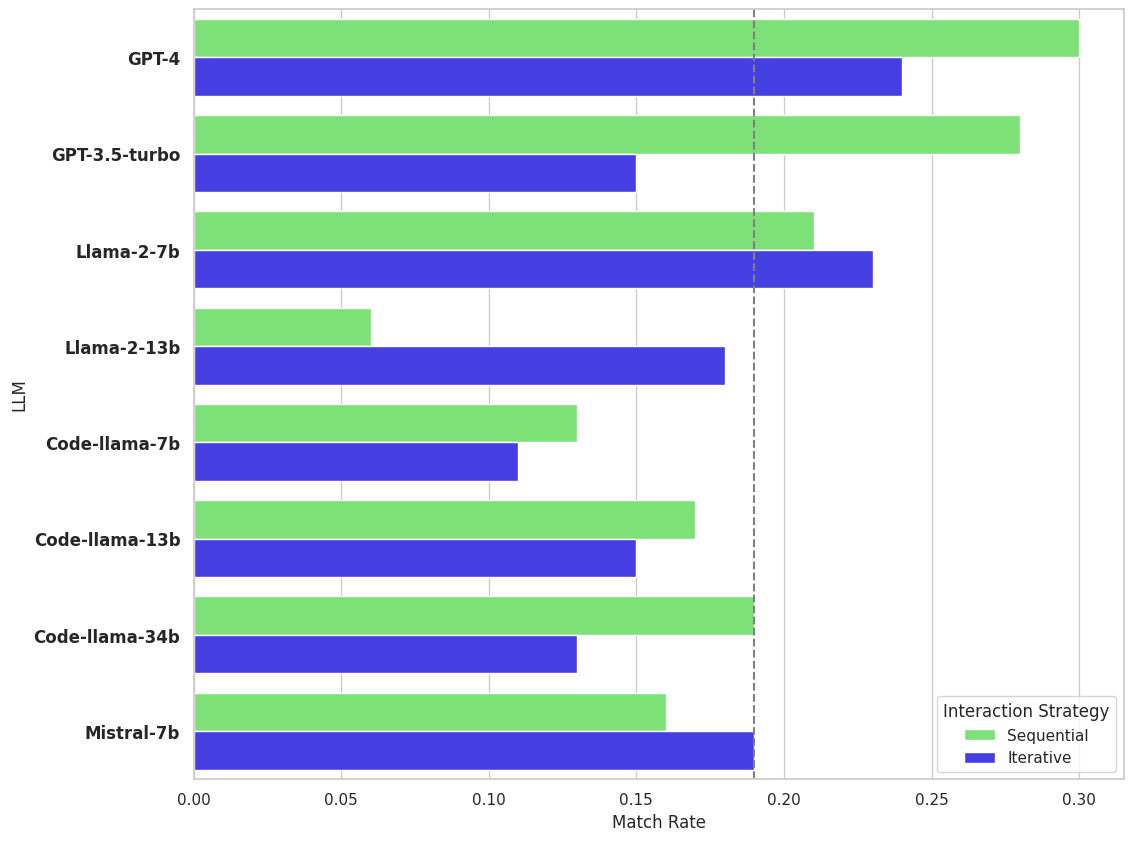}%
}
\caption{Match Rate for Instances with Conclusive Questions}
\label{fig:match-rate-conclusive}
\end{subfigure}
\begin{subfigure}{\columnwidth}
\centering
\resizebox{\columnwidth}{!}{%
\includegraphics[width=\textwidth]{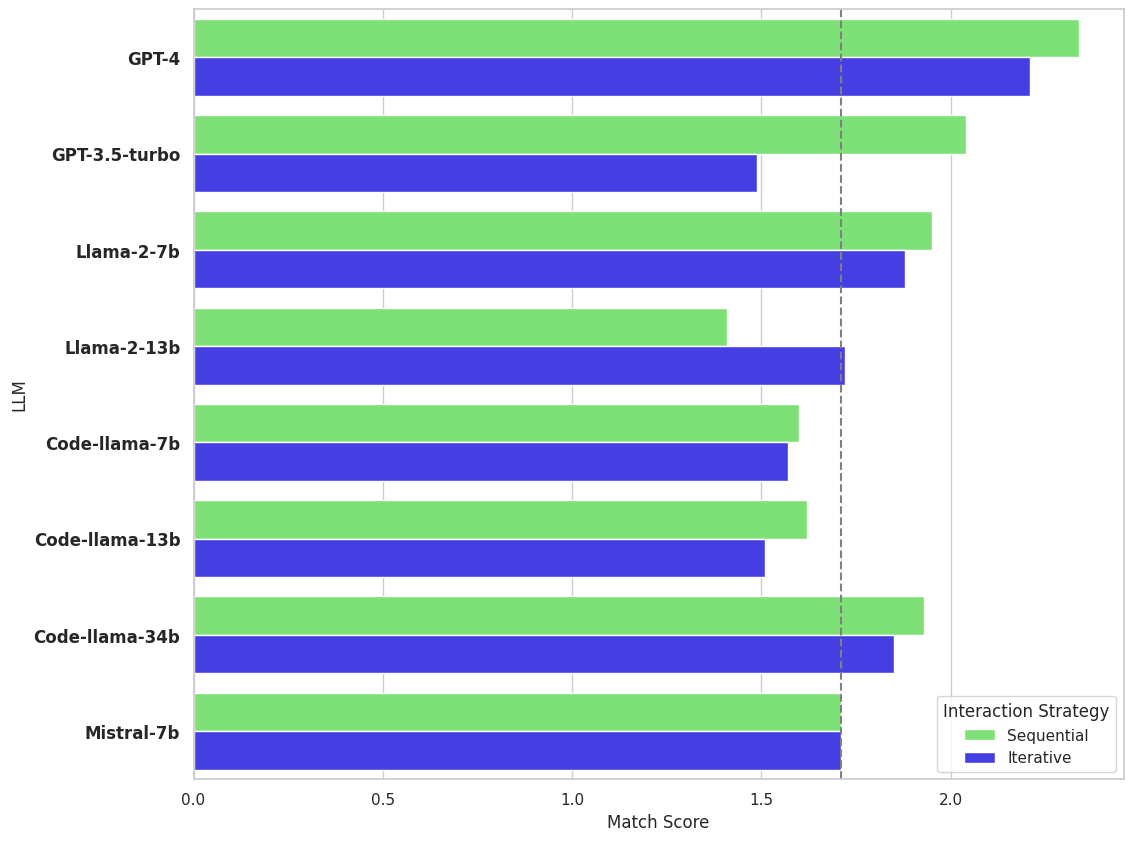}%
}
\caption{Match Score for Instances with Interpretive Questions}
\label{fig:match-score-interpretive}
\end{subfigure}
\caption{Reference-based evaluation results across various interaction strategies and LLMs, with a vertical line representing the performance achieved by a non-interactive LLM agent lacking database context, serving as the baseline for guessing.}
\label{fig:ref-based-result}
\end{figure*}
\begin{figure}[t]
\begin{subfigure}{\columnwidth}
\includegraphics[width=\columnwidth]{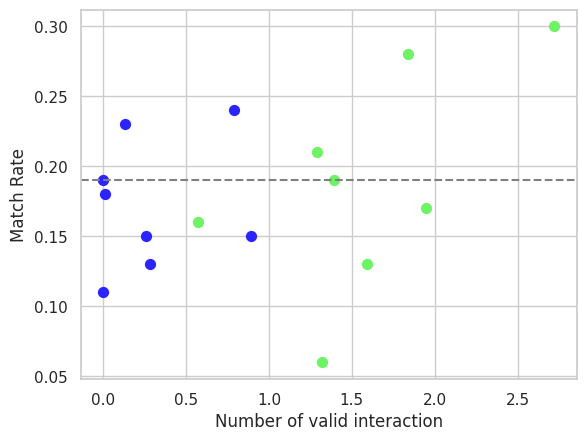}
\caption{Match Rate vs. Valid Interaction for Instances with Conclusive Questions}
\label{fig:match-rate-vs-valid-interaction-conclusive}
\end{subfigure}
\begin{subfigure}{\columnwidth}
\includegraphics[width=\columnwidth]{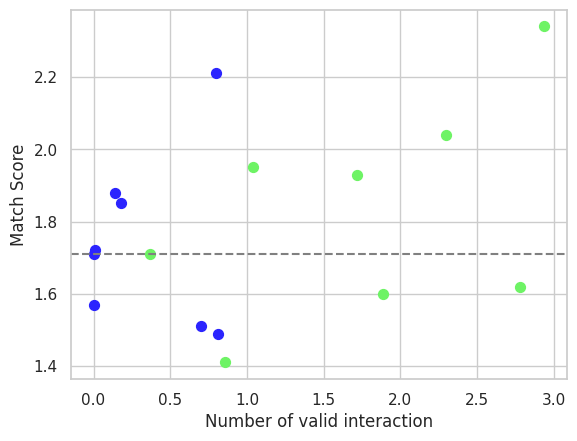}
\caption{Match Score vs. Valid Interaction for Instances with Interpretive Questions}
\label{fig:match-score-vs-valid-interaction-interpretive}
\end{subfigure}
\caption{Correlation between answer quality and number of valid interaction (SQL queries that returned non-empty results)}
\label{fig:performance-vs-valid-interaction}
\end{figure}
\begin{figure}
    \centering
    \includegraphics[width=\columnwidth]{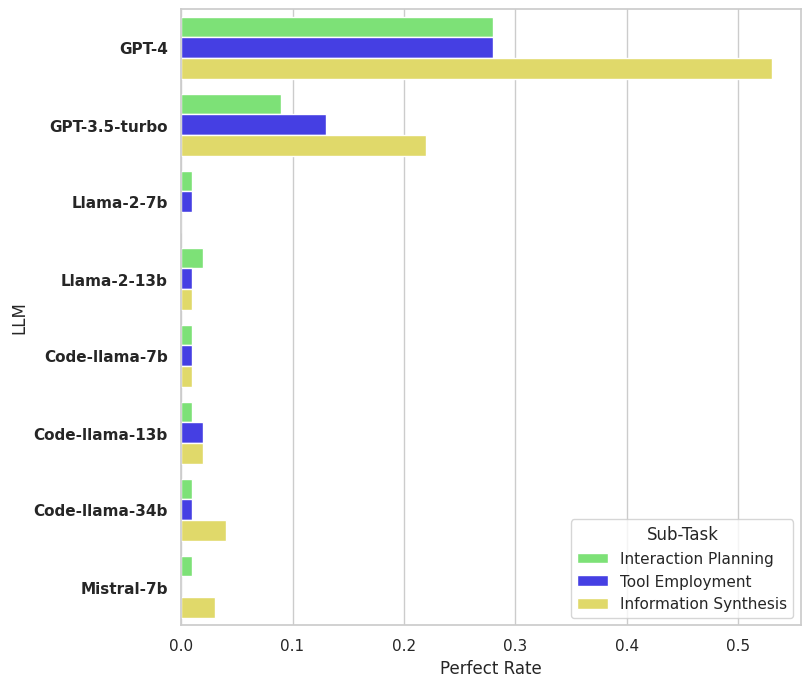}
    \caption{Reference-free multi-agent evaluation results for different sub-tasks and LLMs, all employing sequential interaction strategy.}
    \label{fig:ref-free-result}
\end{figure}

\subsection{Results}
The experimental outcomes are presented in Figures \ref{fig:ref-based-result}, \ref{fig:performance-vs-valid-interaction}, \ref{fig:ref-free-result} and Tables \ref{tab:ref-based-full-result}, \ref{tab:with-without-tool-results}, \ref{tab:ref-free-full-result}, addressing the main research questions posed by our study.

\paragraph{LLM Agent Performance on Proposed Task}
Examining Figures \ref{fig:ref-based-result}, \ref{fig:ref-free-result}, and Tables \ref{tab:ref-based-full-result}, \ref{tab:ref-free-full-result}, it becomes evident that even the state-of-the-art \texttt{GPT-4}, when utilizing a better interaction strategy, correctly answers only 30\% of conclusive questions and achieves an average score of 2.34 on a 1-5 scale. The multi-agent evaluation echoes this, with approximately 30\% of instances deemed perfect, indicating considerable room for improvement.

\paragraph{Improvement from SQL Module Interaction}
Table \ref{tab:with-without-tool-results} compares LLM agents' performances with and without SQL module interaction. A significant improvement is observed in LLMs' performance on conclusive questions when augmented with SQL modules. However, this is not the case for interpretive questions. It is important to note that this result applies to instances with small databases where complete records fit within the context window provided to non-interactive LLMs. This suggests that direct interaction with external modules is particularly beneficial for tasks that demand high precision in retrieval, reinforcing our arguments of LLM limitations in Section \ref{sec:intro}.

\paragraph{The Impact of Interaction Strategies}
Figure \ref{fig:ref-based-result} reveals that, generally, a sequential strategy yields better results across most LLMs, with iterative strategies favoring \texttt{Llama-2-13b} and \texttt{Mistral-7b}. We investigate this trend by analyzing the length of plans, the number of generated SQL queries, and their validity. Notably, when employing iterative strategies, both \texttt{Llama-2-13b} and \texttt{Mistral-7b} engage minimally with the SQL module, with the \texttt{Mistral-7b} agent does not engage with the database at all (empty plan), indicating that it essentially guesses the answers. We set the performance of \texttt{Mistral-7b} with iterative strategy as a reference point for guess-based answers, marked by vertical lines in the figures, and make a note that performances close to this baseline likely result from guesswork. Additionally, we can also notice that agents using iterative strategies tend to plan less and interact minimally with SQL modules, contributing to their underperformance compared to sequential strategies.

\paragraph{Barriers to Effective Interaction}
Figures \ref{fig:ref-free-result} and Table \ref{tab:ref-free-full-result} highlight that planning and tool employment (i.e. SQL generation) are the main hurdles preventing agents from performing well on the proposed tasks. Conversely, agents generally excel at synthesizing retrieved information to produce an accurate and comprehensive answer. This indicates that eliciting better interaction planning via more effective prompting and enhancing LLMs' ability to generate multiple SQL queries in parallel from extended text descriptions are promising areas for future research.

\paragraph{Generalizability Across Different LLMs}
The conclusions regarding interaction strategy appear consistent across various LLMs. However, our hypothesis of interaction barriers primarily holds for the more capable proprietary LLMs like \texttt{GPT-4} and \texttt{GPT-3.5-turbo}, and to a lesser extent, \texttt{Code-llama-34b}. The remaining LLMs did not engage in any meaningful interaction, thus no discernible patterns were noted.

\paragraph{Interaction Depth and Answer Quality}
Figure \ref{fig:performance-vs-valid-interaction} suggests a weak correlation between the number of valid interactions (i.e., agent-generated SQL queries that yield non-empty results) and performance, hinting that more successful retrieval aids in generating more precise and comprehensive answers.

\paragraph{Diversity and Consensus in Multi-agent Evaluation}
Aggregating multiple diverse evaluations and meta-evaluations from LLMs appears to decrease result variance, as evidenced by the increased consensus among meta-reviewer LLMs compared to reviewer LLMs shown in Table \ref{tab:ref-free-full-result}. The scores from meta-reviewers are consistently lower than those from reviewers, indicating that the meta-review process critically considers the issues highlighted by reviewers. This multi-tiered review mechanism ensures that our evaluation framework effectively balances both precision and recall.

\section{Related Work}
\subsection{Augmented Language Models}
The use of external tools to augment language model outputs and mitigate model hallucination has been previously studied in other domains and tasks \cite{mialon2023augmented}. Models augmented with tools like internet search \cite{lazaridou2022internetaugmented}, Python interpreters \cite{gao2023pal}, math equation-generating models \cite{imani2023mathprompter}, and question-answering models \cite{guu2020realm} have empirically shown improvements in accuracy in comparison to their counterpart baseline models. Other models like TALM \cite{parisi2022talm} and Toolformer \cite{schick2023toolformer} for question answering and ToolWriter \cite{gemmell2023generate} for tabular question answering have built on top of these to limit reliance on humans to select tools for question answering models by fine-tuning the models to learn how and when to use tools. Our work differs from these previous works in that we augment language models to use tools in the data-to-text generation domain specifically where the model is expected to not only query from a database with the use of external tools, but also aggregate these results and interpret the data to produce a paragraph-length response to a not necessarily close-ended question.

\subsection{Reasoning and Action}
Our work also draws on elements of frameworks that either prompt the model to repeatedly reason, act upon the reasoning, and update the action plan until the answer is found \cite{yao2023react} or plan out the different components needed to answer a question before retrieving and generating the answer \cite{su-etal-2021-plan-generate}. Other frameworks have used additional language models as the planner to aggregate information retrieved by a diverse inventory of tools \cite{lu2023chameleon}. Our work builds upon some of these frameworks and investigates these in context of the task of long-form data-to-text generation.

\subsection{Text-to-SQL}
The field of Text-to-SQL has been extensively studied as the standard method for database question answering, with significant contributions from a range of studies \citep[\textit{inter alia}]{berant-etal-2013-semantic, zhong-etal-Seq2SQL-2017, yu-etal-2018-spider, yin-neubig-2018-tranx, yu-etal-2019-cosql, wang-etal-2020-rat, yin-etal-2020-tabert, scholak-etal-2021-picard, ren-etal-2021-lego, xie-etal-2022-unifiedskg, cheng-etal-2023-binding, nan-etal-2023-enhancing}. Despite this, our study innovates by introducing a question-answering task over databases that demands the generation of multiple SQL queries to formulate a single answer.

\subsection{Text Generation Evaluation}
Development in automatic evaluation metrics have emerged, utilizing LLMs to evaluate the quality of generated texts \cite{fu2023gptscore, liu2023geval}. These methods have also been adapted for evaluating text pertaining to tabular data \cite{rebuffel2021dataquesteval} and hallucination detection \cite{manakul2023selfcheckgpt}. \cite{wang2023selfconsistency} introduced self-consistency sampling, which has been shown to improve the reasoning performance of the system. This approach involves generating a set of diverse answers and selecting the most common one through majority vote. In our study, we propose a reference-free multi-agent evaluation framework that synthesizes these ideas.

\section{Conclusion}
In conclusion, our investigation reveals the current limitations of LLMs in complex retrieval and reasoning tasks. Augmentation with a SQL module proved beneficial, particularly for conclusive questions, and pointed to the necessity of strategic interaction planning and proficient tool employment. Our findings stress the need for improvement in these areas to enhance LLM effectiveness. Despite the challenge of varying performance across different models, our multi-agent evaluation framework provides a scalable and rigorous method for assessing agent capabilities. We hope that our proposed task and findings will encourage further investigations in LLMs' capabilities of interacting with external modules, inching towards LLMs capable of handling complex tasks with enhanced precision.

\section*{Limitations}

This study acknowledges several constraints that much be considered when interpreting the results. First, our evaluation dataset is relatively small, a limitation primarily due to budget constraints. We plan to expand our dataset to enhance the statistical significance of our findings. Second, our current method does not include a rigorous human evaluation to ascertain the alignment between our automatic evaluation methods and human judgment. We recognize this as a pivotal area for further research and intend to incorporate human evaluation in future. Third, while this study concentrated on tasks that require intensive action and reasoning capabilities, there is room to explore how LLM agents would perform with external modules on similar tasks with less stringent requirements. Fourth, this study's investigation is limited to specific modules, leaving the examination of a broader spectrum of modules unaddressed. Expanding our research to include a more diverse set of modules is a direction we plan to explore in our future work. Lastly, it is important to acknowledge a potential bias in our evaluation methodology stemming from the exclusive use of \texttt{GPT-4} for generating reference answers as well as for evaluating system-generated responses. This reliance could skew the evaluation in favor of \texttt{GPT-4} agent's answers.

\bibliography{anthology,custom}

\begin{thebibliography}{47}
\expandafter\ifx\csname natexlab\endcsname\relax\def\natexlab#1{#1}\fi

\bibitem[{BehnamGhader et~al.(2023)BehnamGhader, Miret, and Reddy}]{behnamghader2023retrieveraugmented}
Parishad BehnamGhader, Santiago Miret, and Siva Reddy. 2023.
\newblock \href {http://arxiv.org/abs/2212.09146} {Can retriever-augmented language models reason? the blame game between the retriever and the language model}.

\bibitem[{Berant et~al.(2013)Berant, Chou, Frostig, and Liang}]{berant-etal-2013-semantic}
Jonathan Berant, Andrew Chou, Roy Frostig, and Percy Liang. 2013.
\newblock \href {https://aclanthology.org/D13-1160} {Semantic parsing on {F}reebase from question-answer pairs}.
\newblock In \emph{Proceedings of the 2013 Conference on Empirical Methods in Natural Language Processing}, pages 1533--1544, Seattle, Washington, USA. Association for Computational Linguistics.

\bibitem[{Brown et~al.(2020)Brown, Mann, Ryder, Subbiah, Kaplan, Dhariwal, Neelakantan, Shyam, Sastry, Askell, Agarwal, Herbert-Voss, Krueger, Henighan, Child, Ramesh, Ziegler, Wu, Winter, Hesse, Chen, Sigler, Litwin, Gray, Chess, Clark, Berner, McCandlish, Radford, Sutskever, and Amodei}]{NEURIPS2020_1457c0d6}
Tom Brown, Benjamin Mann, Nick Ryder, Melanie Subbiah, Jared~D Kaplan, Prafulla Dhariwal, Arvind Neelakantan, Pranav Shyam, Girish Sastry, Amanda Askell, Sandhini Agarwal, Ariel Herbert-Voss, Gretchen Krueger, Tom Henighan, Rewon Child, Aditya Ramesh, Daniel Ziegler, Jeffrey Wu, Clemens Winter, Chris Hesse, Mark Chen, Eric Sigler, Mateusz Litwin, Scott Gray, Benjamin Chess, Jack Clark, Christopher Berner, Sam McCandlish, Alec Radford, Ilya Sutskever, and Dario Amodei. 2020.
\newblock \href {https://proceedings.neurips.cc/paper_files/paper/2020/file/1457c0d6bfcb4967418bfb8ac142f64a-Paper.pdf} {Language models are few-shot learners}.
\newblock In \emph{Advances in Neural Information Processing Systems}, volume~33, pages 1877--1901. Curran Associates, Inc.

\bibitem[{Cheng et~al.(2023)Cheng, Xie, Shi, Li, Nadkarni, Hu, Xiong, Radev, Ostendorf, Zettlemoyer, Smith, and Yu}]{cheng-etal-2023-binding}
Zhoujun Cheng, Tianbao Xie, Peng Shi, Chengzu Li, Rahul Nadkarni, Yushi Hu, Caiming Xiong, Dragomir Radev, Mari Ostendorf, Luke Zettlemoyer, Noah~A. Smith, and Tao Yu. 2023.
\newblock \href {https://openreview.net/forum?id=lH1PV42cbF} {Binding language models in symbolic languages}.
\newblock In \emph{The Eleventh International Conference on Learning Representations}.

\bibitem[{Chowdhery et~al.(2022)Chowdhery, Narang, Devlin, Bosma, Mishra, Roberts, Barham, Chung, Sutton, Gehrmann, Schuh, Shi, Tsvyashchenko, Maynez, Rao, Barnes, Tay, Shazeer, Prabhakaran, Reif, Du, Hutchinson, Pope, Bradbury, Austin, Isard, Gur-Ari, Yin, Duke, Levskaya, Ghemawat, Dev, Michalewski, Garcia, Misra, Robinson, Fedus, Zhou, Ippolito, Luan, Lim, Zoph, Spiridonov, Sepassi, Dohan, Agrawal, Omernick, Dai, Pillai, Pellat, Lewkowycz, Moreira, Child, Polozov, Lee, Zhou, Wang, Saeta, Diaz, Firat, Catasta, Wei, Meier-Hellstern, Eck, Dean, Petrov, and Fiedel}]{chowdhery2022palm}
Aakanksha Chowdhery, Sharan Narang, Jacob Devlin, Maarten Bosma, Gaurav Mishra, Adam Roberts, Paul Barham, Hyung~Won Chung, Charles Sutton, Sebastian Gehrmann, Parker Schuh, Kensen Shi, Sasha Tsvyashchenko, Joshua Maynez, Abhishek Rao, Parker Barnes, Yi~Tay, Noam Shazeer, Vinodkumar Prabhakaran, Emily Reif, Nan Du, Ben Hutchinson, Reiner Pope, James Bradbury, Jacob Austin, Michael Isard, Guy Gur-Ari, Pengcheng Yin, Toju Duke, Anselm Levskaya, Sanjay Ghemawat, Sunipa Dev, Henryk Michalewski, Xavier Garcia, Vedant Misra, Kevin Robinson, Liam Fedus, Denny Zhou, Daphne Ippolito, David Luan, Hyeontaek Lim, Barret Zoph, Alexander Spiridonov, Ryan Sepassi, David Dohan, Shivani Agrawal, Mark Omernick, Andrew~M. Dai, Thanumalayan~Sankaranarayana Pillai, Marie Pellat, Aitor Lewkowycz, Erica Moreira, Rewon Child, Oleksandr Polozov, Katherine Lee, Zongwei Zhou, Xuezhi Wang, Brennan Saeta, Mark Diaz, Orhan Firat, Michele Catasta, Jason Wei, Kathy Meier-Hellstern, Douglas Eck, Jeff Dean, Slav Petrov, and Noah Fiedel. 2022.
\newblock \href {http://arxiv.org/abs/2204.02311} {Palm: Scaling language modeling with pathways}.

\bibitem[{Devlin et~al.(2019)Devlin, Chang, Lee, and Toutanova}]{devlin-etal-2019-bert}
Jacob Devlin, Ming-Wei Chang, Kenton Lee, and Kristina Toutanova. 2019.
\newblock \href {https://doi.org/10.18653/v1/N19-1423} {{BERT}: Pre-training of deep bidirectional transformers for language understanding}.
\newblock In \emph{Proceedings of the 2019 Conference of the North {A}merican Chapter of the Association for Computational Linguistics: Human Language Technologies, Volume 1 (Long and Short Papers)}, pages 4171--4186, Minneapolis, Minnesota. Association for Computational Linguistics.

\bibitem[{Fu et~al.(2023)Fu, Ng, Jiang, and Liu}]{fu2023gptscore}
Jinlan Fu, See-Kiong Ng, Zhengbao Jiang, and Pengfei Liu. 2023.
\newblock \href {http://arxiv.org/abs/2302.04166} {Gptscore: Evaluate as you desire}.

\bibitem[{Gao et~al.(2023{\natexlab{a}})Gao, Madaan, Zhou, Alon, Liu, Yang, Callan, and Neubig}]{pmlr-v202-gao23f}
Luyu Gao, Aman Madaan, Shuyan Zhou, Uri Alon, Pengfei Liu, Yiming Yang, Jamie Callan, and Graham Neubig. 2023{\natexlab{a}}.
\newblock \href {https://proceedings.mlr.press/v202/gao23f.html} {{PAL}: Program-aided language models}.
\newblock In \emph{Proceedings of the 40th International Conference on Machine Learning}, volume 202 of \emph{Proceedings of Machine Learning Research}, pages 10764--10799. PMLR.

\bibitem[{Gao et~al.(2023{\natexlab{b}})Gao, Madaan, Zhou, Alon, Liu, Yang, Callan, and Neubig}]{gao2023pal}
Luyu Gao, Aman Madaan, Shuyan Zhou, Uri Alon, Pengfei Liu, Yiming Yang, Jamie Callan, and Graham Neubig. 2023{\natexlab{b}}.
\newblock \href {http://arxiv.org/abs/2211.10435} {Pal: Program-aided language models}.

\bibitem[{Gemmell and Dalton(2023)}]{gemmell2023generate}
Carlos Gemmell and Jeffrey Dalton. 2023.
\newblock \href {http://arxiv.org/abs/2303.10138} {Generate, transform, answer: Question specific tool synthesis for tabular data}.

\bibitem[{Guu et~al.(2020)Guu, Lee, Tung, Pasupat, and Chang}]{guu2020realm}
Kelvin Guu, Kenton Lee, Zora Tung, Panupong Pasupat, and Ming-Wei Chang. 2020.
\newblock \href {http://arxiv.org/abs/2002.08909} {Realm: Retrieval-augmented language model pre-training}.

\bibitem[{Imani et~al.(2023)Imani, Du, and Shrivastava}]{imani2023mathprompter}
Shima Imani, Liang Du, and Harsh Shrivastava. 2023.
\newblock \href {http://arxiv.org/abs/2303.05398} {Mathprompter: Mathematical reasoning using large language models}.

\bibitem[{Ji et~al.(2023)Ji, Lee, Frieske, Yu, Su, Xu, Ishii, Bang, Madotto, and Fung}]{10.1145/3571730}
Ziwei Ji, Nayeon Lee, Rita Frieske, Tiezheng Yu, Dan Su, Yan Xu, Etsuko Ishii, Ye~Jin Bang, Andrea Madotto, and Pascale Fung. 2023.
\newblock \href {https://doi.org/10.1145/3571730} {Survey of hallucination in natural language generation}.
\newblock \emph{ACM Comput. Surv.}, 55(12).

\bibitem[{Lazaridou et~al.(2022)Lazaridou, Gribovskaya, Stokowiec, and Grigorev}]{lazaridou2022internetaugmented}
Angeliki Lazaridou, Elena Gribovskaya, Wojciech Stokowiec, and Nikolai Grigorev. 2022.
\newblock \href {http://arxiv.org/abs/2203.05115} {Internet-augmented language models through few-shot prompting for open-domain question answering}.

\bibitem[{Lewis et~al.(2020)Lewis, Perez, Piktus, Petroni, Karpukhin, Goyal, K\"{u}ttler, Lewis, Yih, Rockt\"{a}schel, Riedel, and Kiela}]{NEURIPS2020_6b493230}
Patrick Lewis, Ethan Perez, Aleksandra Piktus, Fabio Petroni, Vladimir Karpukhin, Naman Goyal, Heinrich K\"{u}ttler, Mike Lewis, Wen-tau Yih, Tim Rockt\"{a}schel, Sebastian Riedel, and Douwe Kiela. 2020.
\newblock \href {https://proceedings.neurips.cc/paper_files/paper/2020/file/6b493230205f780e1bc26945df7481e5-Paper.pdf} {Retrieval-augmented generation for knowledge-intensive nlp tasks}.
\newblock In \emph{Advances in Neural Information Processing Systems}, volume~33, pages 9459--9474. Curran Associates, Inc.

\bibitem[{Li et~al.(2023)Li, Zhao, Yu, Song, Li, Yu, Li, Huang, and Li}]{li2023apibank}
Minghao Li, Yingxiu Zhao, Bowen Yu, Feifan Song, Hangyu Li, Haiyang Yu, Zhoujun Li, Fei Huang, and Yongbin Li. 2023.
\newblock \href {http://arxiv.org/abs/2304.08244} {Api-bank: A comprehensive benchmark for tool-augmented llms}.

\bibitem[{Liu et~al.(2023)Liu, Iter, Xu, Wang, Xu, and Zhu}]{liu2023geval}
Yang Liu, Dan Iter, Yichong Xu, Shuohang Wang, Ruochen Xu, and Chenguang Zhu. 2023.
\newblock \href {http://arxiv.org/abs/2303.16634} {G-eval: Nlg evaluation using gpt-4 with better human alignment}.

\bibitem[{Lu et~al.(2023)Lu, Peng, Cheng, Galley, Chang, Wu, Zhu, and Gao}]{lu2023chameleon}
Pan Lu, Baolin Peng, Hao Cheng, Michel Galley, Kai-Wei Chang, Ying~Nian Wu, Song-Chun Zhu, and Jianfeng Gao. 2023.
\newblock \href {http://arxiv.org/abs/2304.09842} {Chameleon: Plug-and-play compositional reasoning with large language models}.

\bibitem[{Madaan et~al.(2023)Madaan, Tandon, Gupta, Hallinan, Gao, Wiegreffe, Alon, Dziri, Prabhumoye, Yang, Gupta, Majumder, Hermann, Welleck, Yazdanbakhsh, and Clark}]{madaan2023selfrefine}
Aman Madaan, Niket Tandon, Prakhar Gupta, Skyler Hallinan, Luyu Gao, Sarah Wiegreffe, Uri Alon, Nouha Dziri, Shrimai Prabhumoye, Yiming Yang, Shashank Gupta, Bodhisattwa~Prasad Majumder, Katherine Hermann, Sean Welleck, Amir Yazdanbakhsh, and Peter Clark. 2023.
\newblock \href {http://arxiv.org/abs/2303.17651} {Self-refine: Iterative refinement with self-feedback}.

\bibitem[{Manakul et~al.(2023)Manakul, Liusie, and Gales}]{manakul2023selfcheckgpt}
Potsawee Manakul, Adian Liusie, and Mark J.~F. Gales. 2023.
\newblock \href {http://arxiv.org/abs/2303.08896} {Selfcheckgpt: Zero-resource black-box hallucination detection for generative large language models}.

\bibitem[{Mialon et~al.(2023)Mialon, Dessì, Lomeli, Nalmpantis, Pasunuru, Raileanu, Rozière, Schick, Dwivedi-Yu, Celikyilmaz, Grave, LeCun, and Scialom}]{mialon2023augmented}
Grégoire Mialon, Roberto Dessì, Maria Lomeli, Christoforos Nalmpantis, Ram Pasunuru, Roberta Raileanu, Baptiste Rozière, Timo Schick, Jane Dwivedi-Yu, Asli Celikyilmaz, Edouard Grave, Yann LeCun, and Thomas Scialom. 2023.
\newblock \href {http://arxiv.org/abs/2302.07842} {Augmented language models: a survey}.

\bibitem[{Nakano et~al.(2022)Nakano, Hilton, Balaji, Wu, Ouyang, Kim, Hesse, Jain, Kosaraju, Saunders, Jiang, Cobbe, Eloundou, Krueger, Button, Knight, Chess, and Schulman}]{nakano2022webgpt}
Reiichiro Nakano, Jacob Hilton, Suchir Balaji, Jeff Wu, Long Ouyang, Christina Kim, Christopher Hesse, Shantanu Jain, Vineet Kosaraju, William Saunders, Xu~Jiang, Karl Cobbe, Tyna Eloundou, Gretchen Krueger, Kevin Button, Matthew Knight, Benjamin Chess, and John Schulman. 2022.
\newblock \href {http://arxiv.org/abs/2112.09332} {Webgpt: Browser-assisted question-answering with human feedback}.

\bibitem[{Nan et~al.(2023)Nan, Zhao, Zou, Ri, Tae, Zhang, Cohan, and Radev}]{nan-etal-2023-enhancing}
Linyong Nan, Yilun Zhao, Weijin Zou, Narutatsu Ri, Jaesung Tae, Ellen Zhang, Arman Cohan, and Dragomir Radev. 2023.
\newblock \href {http://arxiv.org/abs/2305.12586} {Enhancing few-shot text-to-sql capabilities of large language models: A study on prompt design strategies}.

\bibitem[{OpenAI(2023)}]{openai2023gpt4}
OpenAI. 2023.
\newblock \href {http://arxiv.org/abs/2303.08774} {Gpt-4 technical report}.

\bibitem[{Parisi et~al.(2022)Parisi, Zhao, and Fiedel}]{parisi2022talm}
Aaron Parisi, Yao Zhao, and Noah Fiedel. 2022.
\newblock \href {http://arxiv.org/abs/2205.12255} {Talm: Tool augmented language models}.

\bibitem[{Paul et~al.(2023)Paul, Ismayilzada, Peyrard, Borges, Bosselut, West, and Faltings}]{paul2023refiner}
Debjit Paul, Mete Ismayilzada, Maxime Peyrard, Beatriz Borges, Antoine Bosselut, Robert West, and Boi Faltings. 2023.
\newblock \href {http://arxiv.org/abs/2304.01904} {Refiner: Reasoning feedback on intermediate representations}.

\bibitem[{Press et~al.(2023)Press, Zhang, Min, Schmidt, Smith, and Lewis}]{press2023measuring}
Ofir Press, Muru Zhang, Sewon Min, Ludwig Schmidt, Noah~A. Smith, and Mike Lewis. 2023.
\newblock \href {http://arxiv.org/abs/2210.03350} {Measuring and narrowing the compositionality gap in language models}.

\bibitem[{Radford et~al.(2019)Radford, Wu, Child, Luan, Amodei, and Sutskever}]{radford2019language}
Alec Radford, Jeff Wu, Rewon Child, David Luan, Dario Amodei, and Ilya Sutskever. 2019.
\newblock Language models are unsupervised multitask learners.

\bibitem[{Rebuffel et~al.(2021)Rebuffel, Scialom, Soulier, Piwowarski, Lamprier, Staiano, Scoutheeten, and Gallinari}]{rebuffel2021dataquesteval}
Clément Rebuffel, Thomas Scialom, Laure Soulier, Benjamin Piwowarski, Sylvain Lamprier, Jacopo Staiano, Geoffrey Scoutheeten, and Patrick Gallinari. 2021.
\newblock \href {http://arxiv.org/abs/2104.07555} {Data-questeval: A referenceless metric for data-to-text semantic evaluation}.

\bibitem[{Ren et~al.(2021)Ren, Dai, Dai, Chen, Yasunaga, Sun, Schuurmans, Leskovec, and Zhou}]{ren-etal-2021-lego}
Hongyu Ren, Hanjun Dai, Bo~Dai, Xinyun Chen, Michihiro Yasunaga, Haitian Sun, Dale Schuurmans, Jure Leskovec, and Denny Zhou. 2021.
\newblock \href {https://proceedings.mlr.press/v139/ren21a.html} {Lego: Latent execution-guided reasoning for multi-hop question answering on knowledge graphs}.
\newblock In \emph{Proceedings of the 38th International Conference on Machine Learning}, volume 139 of \emph{Proceedings of Machine Learning Research}, pages 8959--8970. PMLR.

\bibitem[{Schick et~al.(2023)Schick, Dwivedi-Yu, Dessì, Raileanu, Lomeli, Zettlemoyer, Cancedda, and Scialom}]{schick2023toolformer}
Timo Schick, Jane Dwivedi-Yu, Roberto Dessì, Roberta Raileanu, Maria Lomeli, Luke Zettlemoyer, Nicola Cancedda, and Thomas Scialom. 2023.
\newblock \href {http://arxiv.org/abs/2302.04761} {Toolformer: Language models can teach themselves to use tools}.

\bibitem[{Scholak et~al.(2021)Scholak, Schucher, and Bahdanau}]{scholak-etal-2021-picard}
Torsten Scholak, Nathan Schucher, and Dzmitry Bahdanau. 2021.
\newblock \href {https://doi.org/10.18653/v1/2021.emnlp-main.779} {{PICARD}: Parsing incrementally for constrained auto-regressive decoding from language models}.
\newblock In \emph{Proceedings of the 2021 Conference on Empirical Methods in Natural Language Processing}, pages 9895--9901, Online and Punta Cana, Dominican Republic. Association for Computational Linguistics.

\bibitem[{Shinn et~al.(2023)Shinn, Cassano, Berman, Gopinath, Narasimhan, and Yao}]{shinn2023reflexion}
Noah Shinn, Federico Cassano, Edward Berman, Ashwin Gopinath, Karthik Narasimhan, and Shunyu Yao. 2023.
\newblock \href {http://arxiv.org/abs/2303.11366} {Reflexion: Language agents with verbal reinforcement learning}.

\bibitem[{Shuster et~al.(2022)Shuster, Komeili, Adolphs, Roller, Szlam, and Weston}]{shuster2022language}
Kurt Shuster, Mojtaba Komeili, Leonard Adolphs, Stephen Roller, Arthur Szlam, and Jason Weston. 2022.
\newblock \href {http://arxiv.org/abs/2203.13224} {Language models that seek for knowledge: Modular search \& generation for dialogue and prompt completion}.

\bibitem[{Su et~al.(2021)Su, Vandyke, Wang, Fang, and Collier}]{su-etal-2021-plan-generate}
Yixuan Su, David Vandyke, Sihui Wang, Yimai Fang, and Nigel Collier. 2021.
\newblock \href {https://doi.org/10.18653/v1/2021.findings-emnlp.76} {Plan-then-generate: Controlled data-to-text generation via planning}.
\newblock In \emph{Findings of the Association for Computational Linguistics: EMNLP 2021}, pages 895--909, Punta Cana, Dominican Republic. Association for Computational Linguistics.

\bibitem[{Wang et~al.(2020)Wang, Shin, Liu, Polozov, and Richardson}]{wang-etal-2020-rat}
Bailin Wang, Richard Shin, Xiaodong Liu, Oleksandr Polozov, and Matthew Richardson. 2020.
\newblock \href {https://doi.org/10.18653/v1/2020.acl-main.677} {{RAT-SQL}: Relation-aware schema encoding and linking for text-to-{SQL} parsers}.
\newblock In \emph{Proceedings of the 58th Annual Meeting of the Association for Computational Linguistics}, pages 7567--7578, Online. Association for Computational Linguistics.

\bibitem[{Wang et~al.(2023{\natexlab{a}})Wang, Liu, Yue, Tang, Zhang, Jiayang, Yao, Gao, Hu, Qi, Wang, Yang, Wang, Xie, Zhang, and Zhang}]{wang2023survey}
Cunxiang Wang, Xiaoze Liu, Yuanhao Yue, Xiangru Tang, Tianhang Zhang, Cheng Jiayang, Yunzhi Yao, Wenyang Gao, Xuming Hu, Zehan Qi, Yidong Wang, Linyi Yang, Jindong Wang, Xing Xie, Zheng Zhang, and Yue Zhang. 2023{\natexlab{a}}.
\newblock \href {http://arxiv.org/abs/2310.07521} {Survey on factuality in large language models: Knowledge, retrieval and domain-specificity}.

\bibitem[{Wang et~al.(2023{\natexlab{b}})Wang, Wei, Schuurmans, Le, Chi, Narang, Chowdhery, and Zhou}]{wang2023selfconsistency}
Xuezhi Wang, Jason Wei, Dale Schuurmans, Quoc Le, Ed~Chi, Sharan Narang, Aakanksha Chowdhery, and Denny Zhou. 2023{\natexlab{b}}.
\newblock \href {http://arxiv.org/abs/2203.11171} {Self-consistency improves chain of thought reasoning in language models}.

\bibitem[{Xie et~al.(2022)Xie, Wu, Shi, Zhong, Scholak, Yasunaga, Wu, Zhong, Yin, Wang, Zhong, Wang, Li, Boyle, Ni, Yao, Radev, Xiong, Kong, Zhang, Smith, Zettlemoyer, and Yu}]{xie-etal-2022-unifiedskg}
Tianbao Xie, Chen~Henry Wu, Peng Shi, Ruiqi Zhong, Torsten Scholak, Michihiro Yasunaga, Chien-Sheng Wu, Ming Zhong, Pengcheng Yin, Sida~I. Wang, Victor Zhong, Bailin Wang, Chengzu Li, Connor Boyle, Ansong Ni, Ziyu Yao, Dragomir Radev, Caiming Xiong, Lingpeng Kong, Rui Zhang, Noah~A. Smith, Luke Zettlemoyer, and Tao Yu. 2022.
\newblock \href {https://aclanthology.org/2022.emnlp-main.39} {{U}nified{SKG}: Unifying and multi-tasking structured knowledge grounding with text-to-text language models}.
\newblock In \emph{Proceedings of the 2022 Conference on Empirical Methods in Natural Language Processing}, pages 602--631, Abu Dhabi, United Arab Emirates. Association for Computational Linguistics.

\bibitem[{Yao et~al.(2023)Yao, Zhao, Yu, Du, Shafran, Narasimhan, and Cao}]{yao2023react}
Shunyu Yao, Jeffrey Zhao, Dian Yu, Nan Du, Izhak Shafran, Karthik Narasimhan, and Yuan Cao. 2023.
\newblock \href {http://arxiv.org/abs/2210.03629} {React: Synergizing reasoning and acting in language models}.

\bibitem[{Yin and Neubig(2018)}]{yin-neubig-2018-tranx}
Pengcheng Yin and Graham Neubig. 2018.
\newblock \href {https://doi.org/10.18653/v1/D18-2002} {{TRANX}: A transition-based neural abstract syntax parser for semantic parsing and code generation}.
\newblock In \emph{Proceedings of the 2018 Conference on Empirical Methods in Natural Language Processing: System Demonstrations}, pages 7--12, Brussels, Belgium. Association for Computational Linguistics.

\bibitem[{Yin et~al.(2020)Yin, Neubig, Yih, and Riedel}]{yin-etal-2020-tabert}
Pengcheng Yin, Graham Neubig, Wen-tau Yih, and Sebastian Riedel. 2020.
\newblock \href {https://doi.org/10.18653/v1/2020.acl-main.745} {{T}a{BERT}: Pretraining for joint understanding of textual and tabular data}.
\newblock In \emph{Proceedings of the 58th Annual Meeting of the Association for Computational Linguistics}, pages 8413--8426, Online. Association for Computational Linguistics.

\bibitem[{Yoran et~al.(2023)Yoran, Wolfson, Bogin, Katz, Deutch, and Berant}]{yoran2023answering}
Ori Yoran, Tomer Wolfson, Ben Bogin, Uri Katz, Daniel Deutch, and Jonathan Berant. 2023.
\newblock \href {http://arxiv.org/abs/2304.13007} {Answering questions by meta-reasoning over multiple chains of thought}.

\bibitem[{Yu et~al.(2019)Yu, Zhang, Er, Li, Xue, Pang, Lin, Tan, Shi, Li, Jiang, Yasunaga, Shim, Chen, Fabbri, Li, Chen, Zhang, Dixit, Zhang, Xiong, Socher, Lasecki, and Radev}]{yu-etal-2019-cosql}
Tao Yu, Rui Zhang, Heyang Er, Suyi Li, Eric Xue, Bo~Pang, Xi~Victoria Lin, Yi~Chern Tan, Tianze Shi, Zihan Li, Youxuan Jiang, Michihiro Yasunaga, Sungrok Shim, Tao Chen, Alexander Fabbri, Zifan Li, Luyao Chen, Yuwen Zhang, Shreya Dixit, Vincent Zhang, Caiming Xiong, Richard Socher, Walter Lasecki, and Dragomir Radev. 2019.
\newblock \href {https://doi.org/10.18653/v1/D19-1204} {{C}o{SQL}: A conversational text-to-{SQL} challenge towards cross-domain natural language interfaces to databases}.
\newblock In \emph{Proceedings of the 2019 Conference on Empirical Methods in Natural Language Processing and the 9th International Joint Conference on Natural Language Processing (EMNLP-IJCNLP)}, pages 1962--1979, Hong Kong, China. Association for Computational Linguistics.

\bibitem[{Yu et~al.(2018)Yu, Zhang, Yang, Yasunaga, Wang, Li, Ma, Li, Yao, Roman, Zhang, and Radev}]{yu-etal-2018-spider}
Tao Yu, Rui Zhang, Kai Yang, Michihiro Yasunaga, Dongxu Wang, Zifan Li, James Ma, Irene Li, Qingning Yao, Shanelle Roman, Zilin Zhang, and Dragomir Radev. 2018.
\newblock \href {https://doi.org/10.18653/v1/D18-1425} {{S}pider: A large-scale human-labeled dataset for complex and cross-domain semantic parsing and text-to-{SQL} task}.
\newblock In \emph{Proceedings of the 2018 Conference on Empirical Methods in Natural Language Processing}, pages 3911--3921, Brussels, Belgium. Association for Computational Linguistics.

\bibitem[{Zhong et~al.(2017)Zhong, Xiong, and Socher}]{zhong-etal-Seq2SQL-2017}
Victor Zhong, Caiming Xiong, and Richard Socher. 2017.
\newblock Seq2sql: Generating structured queries from natural language using reinforcement learning.
\newblock \emph{CoRR}, abs/1709.00103.

\bibitem[{Zhuang et~al.(2023)Zhuang, Yu, Wang, Sun, and Zhang}]{zhuang2023toolqa}
Yuchen Zhuang, Yue Yu, Kuan Wang, Haotian Sun, and Chao Zhang. 2023.
\newblock \href {http://arxiv.org/abs/2306.13304} {Toolqa: A dataset for llm question answering with external tools}.

\end{thebibliography}

\onecolumn
\appendix

\section*{Appendix}
\label{sec:appendix}

\begin{figure}[ht]
\begin{itemize}
    \item \textbf{Conclusive questions:}
    \begin{enumerate}
        \item Do dual-enrolled students tend to perform better or worse than their peers in the same degree programs?
        \item Analyze the relationship between teachers' experience and their performance based on the grades received in the courses they have taught.
        \item Investigate any correlations between poker players' performance and factors such as nationality, age, and height.
    \end{enumerate}
    
    \item \textbf{Interpretive questions:}
    \begin{enumerate}
        \item Compare the success metrics between French and non-French singers.
        \item Analyze the impact of record companies on the success of orchestras based on their performance ratings and attendance.
        \item Analyze the performance of the TV series by language and country, and identify any notable patterns or trends.
    \end{enumerate}
\end{itemize}
\caption{Examples of Conclusive and Interpretive Questions}
\label{fig:question-type-demo}
\end{figure}

\begin{figure}[ht]
\centering
\begin{subfigure}{0.9\textwidth}

\textbf{Given the following inputs:}

Question: \{question\}

Reference (Gold) Answer: \{gold\_answer\}

System Generated Answer: \{answer\}

\textbf{Evaluation Process:}

Read the gold answer carefully to understand the precise information it conveys.

Examine the system-generated answer to identify the information presented.

Check for the presence of critical information (such as conclusions) from the gold answer in the system-generated answer.

\textbf{Evaluation Criteria:}

The system-generated answer is considered a "Match" if it contains all the critical information from the gold answer. The presence of additional non-contradictory information in the system-generated answer is acceptable, provided that all the information from the gold answer is included.

\textbf{Output Format:}

If the system-generated answer includes all the critical information from the gold answer, the output should be: "Conclusion: Match"

If any critical information from the gold answer is missing or misrepresented in the system-generated answer, the output should be: "Conclusion: Not Match"

Conclusion: 

\caption{Scoring Metrics for Conclusive Questions}
\end{subfigure}

\vspace{4mm}

\begin{subfigure}{0.9\textwidth}
\textbf{Given the following inputs:}

Question: \{question\}

Reference (Gold) Answer: \{gold\_answer\}

System-Generated Answer: \{answer\}

\textbf{Evaluation Process:}

Familiarize yourself with the gold answer to understand the full scope of information it contains.

Analyze the system-generated answer to identify the information that has been captured.

Compare the two answers to determine how much of the gold answer's information is reflected in the system-generated answer.

\textbf{Scoring Metrics:}

Score 1: The system-generated answer lacks almost all the key points that the comprehensive gold answer provides.

Score 2: The system-generated answer includes some key points from the gold answer but misses others, and it may include additional details not found in the gold answer.

Score 3: The system-generated answer captures most of the key information from the gold answer, but there are noticeable omissions or additions.

Score 4: The system-generated answer encompasses all key points from the gold answer and also introduces more information not covered in the gold answer.

Score 5: The system-generated answer perfectly mirrors the gold answer, containing all the information with no omissions or additions.

\textbf{Output Format:}

Provide a score between 1 to 5 based on the evaluation. The output should be: "Score: [1/2/3/4/5]"

Score: 

\caption{Scoring Metrics for Interpretive Questions}
\end{subfigure}

\caption{Prompts used for evaluating system generated answers for conclusive and interpretive questions}
\label{fig:ref-based-rubric}
\end{figure}

\begin{figure}[ht]
\centering
\resizebox{0.75\textwidth}{!}{%
\begin{subfigure}{\textwidth}
\textbf{Problem Context:}\\
A planning agent has been tasked to devise a solution to a user question related to a database. Given the question and the database's description, the agent proposes a plan detailing the type of information it would retrieve from the database to answer the question effectively.\\
\textbf{Your Task:}\\
You are to evaluate the plan's relevance and comprehensiveness. Assess whether the plan can indeed retrieve the necessary information to address the user's question.\\
\textbf{Inputs:}\\
User Question: \\
\{question\}\\
Database Description: \\
\{database\_text\}\\
Agent's Proposed Plan: \\
\{plan\}\\
\textbf{Evaluation Criteria:}\\
Relevance: Does the plan target relevant pieces of information from the database that directly pertain to the user's question?\\
Comprehensiveness: Is the plan exhaustive, ensuring all necessary pieces of information are retrieved to fully answer the user's question?\\
\textbf{Plan Definitions:}\\
Perfect Plan: A plan that is both relevant and comprehensive, ensuring that the user's question can be answered completely without missing any essential data points.\\
Imperfect Plan: A plan that misses out on some relevant information, or includes unnecessary steps, thus not providing a complete or accurate solution to the user's question.\\
\textbf{Response Format:}\\
Rationale: Begin with a detailed explanation of your evaluation. Discuss the strengths or weaknesses of the plan based on the relevance and comprehensiveness criteria.\\
Final Decision: After providing the rationale, conclude with one of the following decisions:\\
- Perfect: If you believe the plan meets both the relevance and comprehensiveness criteria effectively.\\
- Imperfect: If you find the plan lacking in any aspect, be it relevance or comprehensiveness.
\caption{Review Criteria for Interaction Planning}
\end{subfigure}%
}

\vspace{4mm} %

\resizebox{0.75\textwidth}{!}{%
\begin{subfigure}{\textwidth}
\textbf{Problem Context:}\\
As the "editor-in-chief", you are tasked with evaluating the reviews provided by multiple reviewers on a planning agent's proposed plan to answer a database-related user question. Each review contains a detailed rationale and a final decision.\\
\textbf{Your Task:}\\
Your goal is to compare and assess the rationales provided by the reviewers, and then make a final, conclusive decision about the planning agent's proposal. This decision should be based on a comprehensive understanding of the reviewers' perspectives and the evidence they present.\\
\textbf{Inputs:}\\
User Question:\\
\{question\}\\
Database Description:\\
\{database\_text\}\\
Agent's Proposed Plan:\\
\{plan\}\\
Reviewers' Rationales and Decisions:\\
\{IP\_reviews\}\\
\textbf{Evaluation Criteria:}\\
Review Consistency: Are the reviewers' rationales and decisions consistent with each other?\\
Evidence Quality: Is the evidence provided in the rationales substantial and convincing enough to make a definitive conclusion?\\
Final Decision Basis: Does the aggregated perspective of the reviewers lead to a clear final decision?\\
\textbf{Response Format:}\\
Rationale: Begin with a detailed explanation comparing the rationales provided by the reviewers. Highlight consistencies or discrepancies among them and discuss how these influenced your final decision.\\
Final Decision: After providing the rationale, conclude with one of the following decisions:\\
- Perfect: If the aggregated insights from reviewers suggest that the planning agent's proposal is both relevant and comprehensive.\\
- Imperfect: If the combined reviews indicate that the planning agent's proposal is lacking in either relevance or comprehensiveness.
\caption{Meta-Review Criteria for Interaction Planning}
\end{subfigure}%
}
\caption{Prompts used for reviewing and meta-reviewing interaction planning}
\label{fig:multi-agent-framework-IP-rubric}
\end{figure}

\begin{figure}[ht]
\centering
\resizebox{0.75\textwidth}{!}{%
\begin{subfigure}{\textwidth}
\textbf{Problem Context:}\\
An agent is given a question, a database for retrieving relevant context, and a plan of how to perform the retrieval. It has been tasked to translate the plan into accurate and executable SQL queries. These queries should correspond to the given plan and effectively retrieve the relevant information from the database to address the user's question, adhering to the database structure provided.\\
\textbf{Your Task:}\\
You are to evaluate the correctness and alignment of the SQL queries generated by the agent based on the plan provided. Also, review the execution results to determine if they fulfill the user's requirements as stipulated in the plan.\\
\textbf{Inputs:}\\
User Question: \{question\}\\
Database Description: \{database\_text\}\\
Search Plan: \{plan\}\\
Agent's Proposed SQL Queries and Execution Results: \{sql\_results\}\\
\textbf{Evaluation Criteria:}\\
Correctness: Are the SQL queries syntactically and semantically correct, and do they retrieve the expected data from the database?\\
Alignment: Do the SQL queries align with the steps outlined in the initial plan?\\
Execution Results: Does the outcome of the SQL queries correspond to the desired results based on the user's question and the initial plan?\\
\textbf{Query Definitions:}\\
Perfect Queries: All SQL queries are correct, aligned, and ensure that the user's question is addressed in accordance with the initial plan.\\
Imperfect Queries: There is at least one SQL query that has errors, misalignments, or does not produce the expected results as outlined in the initial plan.\\
\textbf{Response Format:}\\
Rationale: Begin with a detailed explanation of your evaluation. Address the SQL queries' correctness, their alignment with the initial plan, and the resulting output's relevance to the user's query.\\
Final Decision: After providing the rationale, conclude with one of the following decisions:\\
- Perfect: If all SQL queries are correct, aligned with the plan, and the results answer the user's question as expected.\\
- Imperfect: If you find any discrepancies in correctness, alignment, or the execution results of the proposed SQL queries.

\caption{Review Criteria for Tool Employment}
\end{subfigure}%
}

\vspace{4mm} %

\resizebox{0.75\textwidth}{!}{%
\begin{subfigure}{\textwidth}
\textbf{Problem Context:}\\
As the "editor-in-chief", you are presented with multiple reviews evaluating an agent's capability to generate SQL queries from a given plan to answer a user question using a specified database. Each review contains an in-depth rationale and a final decision regarding the correctness, alignment, and execution results of the SQL queries.\\
\textbf{Your Task:}\\
Your goal is to compare and assess the rationales provided by the reviewers, weighing their evidence and perspectives, and then make a final, conclusive decision regarding the agent's SQL queries based on the aggregated reviews.\\
\textbf{Inputs:}\\
User Question: \{question\}\\
Database Description: \{database\_text\}\\
Search Plan: \{plan\}\\
Agent's Proposed SQL Queries and Execution Results: \{sql\_results\}\\
Reviewers' Rationales and Decisions: \{TE\_reviews\}\\
\textbf{Evaluation Criteria:}\\
Review Consistency: Do the reviewers agree in their evaluations, or are there conflicting perspectives?\\
Evidence Quality: Are the rationales provided by reviewers substantial and convincing?\\
Final Decision Basis: Based on the aggregated insights of the reviewers, is there a clear and justifiable final decision?\\
\textbf{Response Format:}\\
Rationale: Begin with a detailed comparison of the rationales provided by the reviewers. Address any consistencies or discrepancies in their evaluations, emphasizing how these observations influenced your final decision.\\
Final Decision: After analyzing the rationales, conclude with one of the following decisions:\\
- Perfect: If the collective insights suggest that all the agent's SQL queries are accurate, aligned, and answer the user's question as stipulated.\\
- Imperfect: If the combined reviews reveal issues in correctness, alignment, or the execution results of the agent's SQL queries.

\caption{Meta-Review Criteria for Tool Employment}
\end{subfigure}%
}

\caption{Prompts used for reviewing and meta-reviewing tool employment}
\label{fig:multi-agent-framework-TE-rubric}
\end{figure}

\begin{figure}[ht]
\centering
\resizebox{0.66\textwidth}{!}{%
\begin{subfigure}{\textwidth}
\textbf{Problem Context:}\\
An agent is presented with a user's question, a plan to extract more context for answering the question, and a search history containing SQL queries used to retrieve this context from the database. The agent's task is to synthesize all the given information to construct a coherent answer to the question.\\
\textbf{Your Task:}\\
You are to evaluate the synthesis produced by the agent. Assess whether the agent's response accurately interprets the SQL queries and their execution results. Furthermore, determine if the synthesized answer addresses the user's question both correctly and comprehensively.\\
\textbf{Inputs:}\\
User Question:\\
\{question\}\\
Database Description:\\
\{database\_text\}\\
Search Plan:\\
\{plan\}\\
SQL Queries and Execution Results:\\
\{sql\_results\}\\
Agent's Synthesized Answer:\\
\{answer\}\\
\textbf{Evaluation Criteria:}\\
Interpretation Accuracy: Does the agent's answer demonstrate a correct understanding of the SQL queries and their execution results?\\
Answer Correctness: Is the agent's synthesized answer accurate in terms of the given information?\\
Comprehensiveness: Does the agent's answer cover all aspects of the user's question based on the context retrieved?\\
\textbf{Answer Definitions:}\\
Perfect Answer: An answer that accurately interprets the SQL queries and results, and addresses the user's question both correctly and comprehensively.\\
Imperfect Answer: An answer that either misinterprets the SQL information, or does not completely and accurately address the user's question.\\
\textbf{Response Format:}\\
Rationale: Begin with a detailed explanation of your evaluation. Discuss the strengths or weaknesses of the agent's synthesized answer based on the criteria of interpretation accuracy, correctness, and comprehensiveness.\\
Final Decision: After providing the rationale, conclude with one of the following decisions:\\
- Perfect: If you believe the agent's answer meets all evaluation criteria effectively.\\
- Imperfect: If you identify any shortcomings in interpretation accuracy, correctness, or comprehensiveness of the answer.

\caption{Review Criteria for Information Synthesis}
\end{subfigure}%
}

\vspace{4mm} %

\resizebox{0.66\textwidth}{!}{%
\begin{subfigure}{\textwidth}
\textbf{Problem Context:}\\
As the "editor-in-chief", you are tasked with evaluating multiple reviews that assess an agent's synthesis of an answer based on a user's question, a search plan, and the results of executed SQL queries. Each review contains a detailed rationale and a final decision on the agent's capability to coherently integrate the information and answer the user's question.\\
\textbf{Your Task:}\\
Your role is to compare and evaluate the rationales provided by the reviewers, integrating their insights and perspectives. Based on this aggregated understanding, make a final, conclusive decision about the agent's synthesized answer.\\
\textbf{Inputs:}\\
User Question:\\
\{question\}\\
Database Description:\\
\{database\_text\}\\
Search Plan:\\
\{plan\}\\
SQL Queries and Execution Results:\\
\{sql\_results\}\\
Agent's Synthesized Answer:\\
\{answer\}\\
Reviewers' Rationales and Decisions:\\
\{IS\_reviews\}\\
\textbf{Evaluation Criteria:}\\
Review Consistency: Are there shared perspectives among the reviewers, or do they have conflicting views?\\
Evidence Quality: Do the reviewers present substantial and compelling evidence in their rationales?\\
Final Decision Basis: Does the collective insight of the reviewers lead to a clear, definitive conclusion about the agent's answer?\\
\textbf{Response Format:}\\
Rationale: Start with a comprehensive comparison of the rationales given by the reviewers. Address any commonalities or differences in their evaluations and describe how these factors influenced your final decision.\\
Final Decision: After dissecting the reviewers' insights, decide on one of the following:
- Perfect: If the collective evaluations suggest that the agent's synthesized answer meets all the required criteria.\\
- Imperfect: If the integrated reviews indicate issues in the agent's interpretation, correctness, or comprehensiveness.

\caption{Meta-Review Criteria for Information Synthesis}
\end{subfigure}%
}

\caption{Prompts used for reviewing and meta-reviewing information synthesis}
\label{fig:multi-agent-framework-IS-rubric}
\end{figure}

\end{document}